%% file: main.tex
\documentclass[sigconf]{acmart}

\def\BibTeX{{\rm B\kern-.05em{\sc i\kern-.025em b}\kern-.08emT\kern-.1667em\lower.7ex\hbox{E}\kern-.125emX}}
    
\usepackage{adjustbox}
\usepackage{colortbl}
\usepackage{amsmath}
\usepackage{natbib}
\usepackage{tikz}
\usepackage{url}
\usepackage{textcomp}
\DeclareMathOperator*{\argmin}{arg\,min} 
\newcommand{\tcolor}{0.85, 0.85, 0.85}

\let\oldftcp\footnotetextcopyrightpermission
\renewcommand\footnotetextcopyrightpermission[1]{\oldftcp{%
\textcopyright{} 2020, Copyright is with the authors.
Published in the Proceedings of the BDA 2019 Conference
(15-18 October 2019, Lyon, France).
Distribution of this paper is permitted under the terms
of the Creative Commons license CC-by-nc-nd 4.0.\\
\textcopyright{} 2020, Droits restant aux auteurs.
Publi{\'e} dans les actes de la conf{\'e}rence BDA 2019
(15 au 18 octobre 2019, Lyon, France).
Redistribution de cet article autoris{\'e}e selon les termes
de la licence Creative Commons CC-by-nc-nd 4.0.}}

\begin{document}

\title[Patient trajectory prediction in the Mimic-III dataset]{
    Patient trajectory prediction in the\\ Mimic-III dataset, challenges and pitfalls
}

\author{Jose F Rodrigues-Jr}
\email{junio@usp.br}
\orcid{0000-0001-8318-1780}
\affiliation{
  \institution{University of Sao Paulo}
  \city{Sao Carlos, SP, Brazil}
}

\author{Gabriel Spadon}
\email{spadon@usp.br}
\orcid{0000-0001-8437-4349}
\affiliation{
  \institution{University of Sao Paulo}
  \city{Sao Carlos, SP, Brazil}
}

\author{Bruno Brandoli}
\email{brunobrandoli@gmail.com}
\affiliation{
  \institution{Dalhousie University}
  \city{Halifax, NS, Canada}
}

\author{Sihem Amer-Yahia}
\email{sihem.amer-yahia@imag.fr}
\affiliation{
  \institution{Centre National de la Recherche Scientifique}
  \city{Grenoble, France}
}

\begin{abstract}
Automated medical prognosis has gained interest as artificial intelligence evolves and the potential for computer-aided medicine becomes evident. Nevertheless, it is challenging to design an effective system that, given a patient's medical history, is able to predict probable future conditions. Previous works, mostly carried out over private datasets, have tackled the problem by using artificial neural network architectures that cannot deal with low-cardinality datasets, or by means of non-generalizable inference approaches. We introduce a Deep Learning architecture whose design results from an intensive experimental process. The final architecture is based on two parallel Minimal Gated Recurrent Unit networks working in bi-directional manner, which was extensively tested with the open-access Mimic-III dataset. Our results demonstrate significant improvements in automated medical prognosis, as measured with Recall@k. We summarize our experience as a set of relevant insights for the design of Deep Learning architectures. Our work improves the performance of computer-aided medicine and can serve as a guide in designing artificial neural networks used in prediction tasks.
\end{abstract}



\maketitle

\section{Introduction}
\label{sec:intro}
\vspace{0.25cm}
Routinely, health care professionals have to deal with patient records that carry years, even decades, of clinical evidence. It is their job to digest all this information to make the most accurate recommendations regarding the patient's health and the most adequate treatments. Manually processing such amounts of information is time and effort-intensive and physicians can certainly benefit from the aid of an automated prognosis system \cite{articleMiller2017}. The accurate processing of the entire profile of a patient formed by a sequence of events can lead to more precise prognoses, fostering preventive medicine practices, delineating healthier habits, and aiding the health sector as a whole. The same benefits hold for health insurance companies who are interested in predicting the possible outcomes of their clients and proposing fair contracts and conditions.

Automated prognosis can benefit from the wide adoption of Electronic Health Records (EHRs) \cite{Henry2013AdoptionOE}, a practice that is leading to a massive production of computer-ready clinical data. One branch of this research is referred to as {\em Patient Trajectory Prediction, or Disease Progression.} This field relates to taking into account temporally-ordered sets of clinical data and having the computer learn what the next most probable event is. For our specific settings, a clinical event refers to a patient admitted to a hospital; along the course of this admission, a set of diagnostic outcomes are generated and encoded according to the International Statistical Classification of Diseases and Related Health Problems, 9th revision (ICD-9\footnote{\url{https://apps.who.int/iris/handle/10665/39473}}). Given a sequence of admissions referring to a patient, we want to predict the most probable diagnoses that will be observed in that patient's next admission. The task, although modeled for computational processing, is similar to what health professionals repeatedly do when faced with historical clinical profiles to delineate an expected prognosis. For this task, Deep Learning (DL) techniques have increasingly gained interest due to their adequacy in dealing with large amounts of sequential data and due to their convincing results in prediction/classification problems, as we review in Section~\ref{sec:relatedworks}.

Despite the advances in the field, the use of DL for patient trajectory prediction, a modality of sequence-to-sequence prediction that depends of a temporal context, stumbles in several challenges. Initially, after data preparation to filter out inconsistencies, the data must be inspected regarding its distribution, a process that guides the proper encoding of the information and its respective modeling to a tensor representation. The processing, then, depends on the proper definition of a DL architecture taking into account dozens of different kinds of processing units found in the literature -- the ideal method must be capable of dealing with a memory of past events, answering for the temporal aspect of the problem. Furthermore, a myriad of hyperparameters must be taken into account, including the number and size of layers, activation functions, convergence, regularization, loss function, and optimization, among others. In our specific setting, we found that the small cardinality of the data and its highly granular encoding posed strong hurdles that steered all the research process -- our findings orbit all these challenges.

We report on the use of Deep Learning techniques applied to the open-access dataset Medical Information Mart for Intensive Care III (Mimic-III) \cite{Johnson2016} provided by the Massachusetts Institute of Technology. We describe efforts on using this rich dataset to build a medical prognosis method using recurrent artificial neural networks. Our methodology can be briefly summarized as: pre-processing the Mimic-III ICD-9 diagnosis encoding according to a less granular encoding scheme provided by HCUP, the North-American Healthcare Cost and Utilization Project (Section~\ref{sec:mimic}); using two Minimal Gated Recurrent Unit networks organized in a parallel bidirectional structure able to deal with both the low cardinality of Mimic-III and with its temporal nature (Section \ref{sec:LIG-Doctor}); and experimenting with several methods reported in the literature to provide a broad panorama on how to tackle the specific settings of Mimic-III (results reported in Section~\ref{sec:experiments}).
We named our methodology LIG-Doctor after the name of our research group, the {\it Laboratoire d'Informatique de Grenoble}.

We summarize our contributions as follows:
\begin{enumerate}
    \item {\it Elucidation on DL methods}: we extensively tested DL methods frequently referenced in the literature; our results allowed us to identify pitfalls and good choices applicable to Mimic-III and, possibly, in a more general context;
    \item {\it Methodology for computational prognosis}: as a result of extensive experiments with multiple methods, we reached a methodology that demonstrated to be successful for Mimic-III in comparison to other methods considered as  state of the art -- the source code of this project is accessible at GitHub\footnote{\url{https://github.com/jfrjunio/LIG-Doctor}};
    \item {\it Dataset insight}: we discuss the characteristics of Mimic-III with respect to its potential for computational medical prognosis; we focus on aspects of its cardinality, encoding granularity, and clinical aspects (Intensive Care Unit) demonstrating methodological choices that result in more accurate outcomes.
\end{enumerate}

\section{Related works}
\label{sec:relatedworks}
\vspace{0.25cm}
Currently, the comparison of computational medical prognostic methods is hampered by the fact that the works in the literature use different datasets. Each dataset carries specific characteristics, like domain, cardinality, structure, and data encoding -- quite often, the datasets are proprietary and not available for broader use. Unlike most datasets, the open-access Mimic-III dataset is a highly structured and semantically rich dataset that has gained attention within the research community. Despite the difficult to compare existing works, we review important proposals that deal with medical prognosis and highlight the challenges that lie in the field, allowing one to assess this research.

The work of Pham et al. \cite{PHAM2017218}, named DeepCare, uses an LSTM~\cite{HochreiterLSTM} recurrent neural network that, in addition to the codes found in the Electronic Medical Record, concatenates extra features to the data to aid the learning process. The concatenation occurs after an embedding layer and includes intervention codes, elapsed time, and admission method; the authors also use pooling as an attention mechanism to focus on the most important diagnostic inputs of an admission sequence. Similar to DeepCare, we experimented with the use of embedding and extra features, as we report in Section~\ref{sec:experiments}. The use of embedding, though, resulted in a reduced recall in all the cases; the use of extra features, such as type and duration of admission, slightly improved our recall measures. The possible explanation is that Mimic-III is a general clinic dataset whose cases range, for example, from childbirth to myocardial infarction while the datasets used by Pham et al. refer to specific disorders (diabetes and mental health). Therefore, the spectrum of Mimic-III with respect to its features is broader, in which case, extra features did not represent strong information gains. Unfortunately, we were not able to reproduce DeepCare, as the code provided by the authors is incomplete.

Choi et al.~\cite{Choi2016-doctorai} introduce the Doctor AI methodology, which is based on a Gated Recurrent Unit network. Their architecture uses an embedding layer to reduce the dimensionality of the input admission sequences, and an optional SKip-gram representation of the diagnoses codes. They report experiments on a dataset with more than 14 million admissions related to a case-control study relative to heart failure. Their dataset refers to a specific disease and is more than 240 times bigger than Mimic-III. Choi et al. also report that they were able to perform transfer learning using their model to make predictions over Mimic-III; their results were not as precise as those of LIG-Doctor neither with transfer learning, nor when straightly applied over Mimic-III, as we present in Section~\ref{sec:experiments}.

Prognostic medicine has been tackled by other approaches not based on DL. Jensen et al.~\cite{Jensen2014} describe a statistical frequentist inferential technique to characterize trajectories of interest in the population of Denmark. Their approach, although sound, does not adjusts for other settings as it demands customized modeling and formulation. Additionally, it was not designed to learn new trajectories from new data. Some works have used Markovian models to compute the conditional probability $p(H \rightarrow a| H)$, that is, the probability of an admission $a$ given the clinical record $H$. Wang et al.~\cite{Wang:2014:ULD:2623330.2623754}, for example, describes a two-part method that uses Bayesian and Markovian principles for Chronic Obstructive Pulmonary Diseases. Their method primes for being unsupervised and quite precise in its specific disease context. The drawback is that it depends on co-morbidity information provided by specialists, which is hardly ever available in significant numbers. Furthermore, their modeling has a complexity that cannot be disregarded, making it hard to adapt the method to new settings. More generally, Arandjelovic~\cite{10.1093/bioinformatics/btv508} claims and demonstrates that, for trajectory prediction, Markovian models depend on constrained assumptions to reduce the number of possible historical sequences. This fact leads to limited applicability, especially because such models cannot deal with admissions that contrast concerning their severity; that is, a routine admission would simply erase the model's memory of a previous severe condition.

Other approaches rely on Hawkes Processes~\cite{Linderman:2014:DLN:3044805.3045050}, a sort of point process or probabilistic model for random scatterings. Such a method is used for describing the occurrence of events over time, as in the case of patient admissions to a hospital and respective diagnoses. The drawback of such works is that they apply strictly to diseases and not to patients, which severely reduces the applicability of the models. Also, the number of parameters grows quadratically with the number of diseases, incurring in a huge computational cost~\cite{Choi:2015:CDN:2919336.2920580}.

\section{The Mimic-III dataset}
\label{sec:mimic}
\vspace{0.25cm}
This work focuses on the open-access dataset Medical Information Mart for Intensive Care III (Mimic-III)~\cite{Johnson2016} provided by the Massachusetts Institute of Technology. This dataset integrates deidentified, comprehensive clinical data of patients admitted to the critical care unit of the Beth Israel Deaconess Medical Center in Boston, Massachusetts. The access to the dataset is open, but it is conditioned to a strictly controlled user agreement. One of the goals of the Mimic-III effort is to allow the reproduction of clinical studies worldwide, making medical-related research comparable via a standard referential. In fact, during this research, we noticed a flagrant problem; the majority of previous researches rely on private datasets, which prevents reproduction and comparison.

\begin{figure}[ht!]
    \centering
    \includegraphics[width=.95\linewidth]{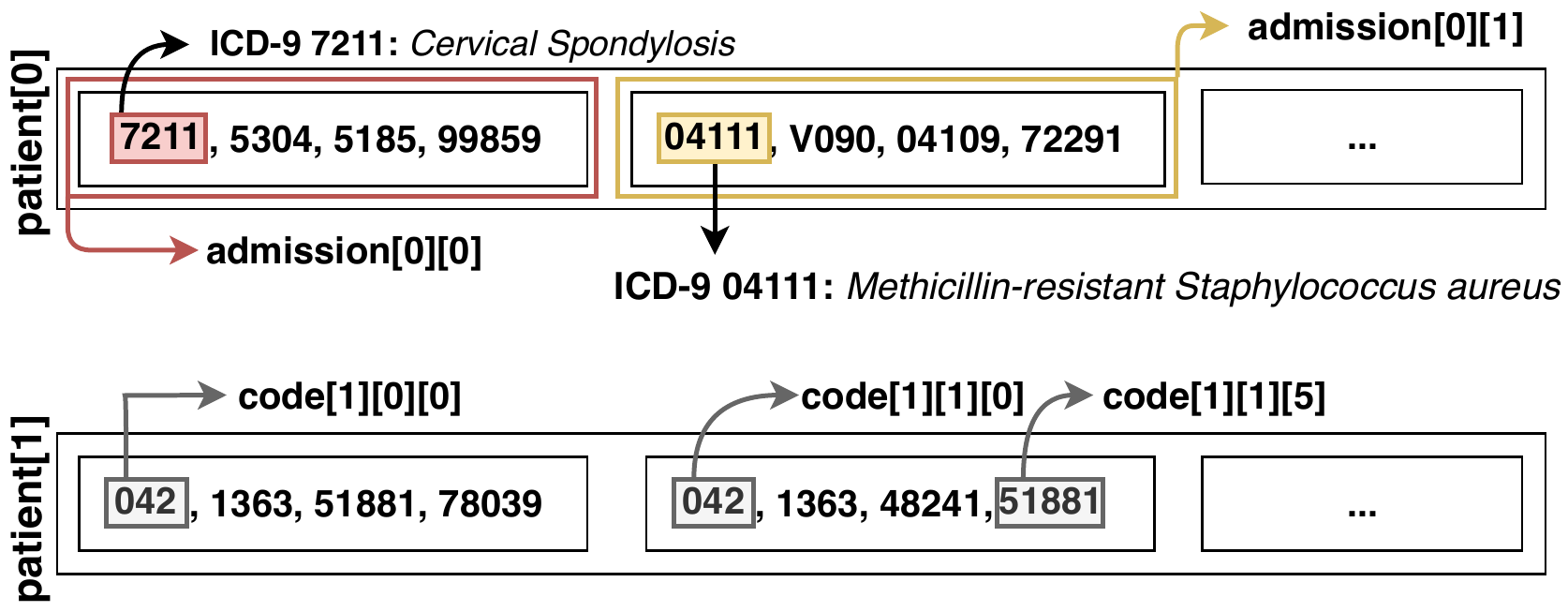}
    \vspace{0.25cm}
    \caption{Three-level record organization extracted from Mimic-III. Many patients with multiple admissions, each one described by a set of ICD-9 codes.}
    \label{fig:LIG-Doctor-File}
\end{figure}

Mimic-III is a well-structured validated dataset with 58,976 admissions from 48,520 patients whose conditions relate to heart, surgical, and trauma conditions, all in demand for critical care. The data is semantically rich including bedside monitoring, laboratory tests, billing, demographics, diagnoses, and procedures. These last two pieces of information are properly encoded using the ICD-9 standard. The dataset has been used for different research purposes such as medication dosing \cite{Ghassemi2014} and mortality prediction \cite{Pirracchio2015}. In this work, we explore the admissions and diagnoses to predict trajectories. A patient's admission refers to a set of diagnostic codes that describe what happened during a hospital stay -- see Figure \ref{fig:LIG-Doctor-File}. 

\subsection{Data issues}
\label{sec:data}
\vspace{0.25cm}
In the task of trajectory prediction, Mimic-III represents a real challenge because its cardinality is relatively small, and because it uses the ICD-9 encoding, which is highly granular.

\subsubsection*{Cardinality}
\begin{figure}[ht!]
    \centering
    \includegraphics[width=0.5\textwidth]{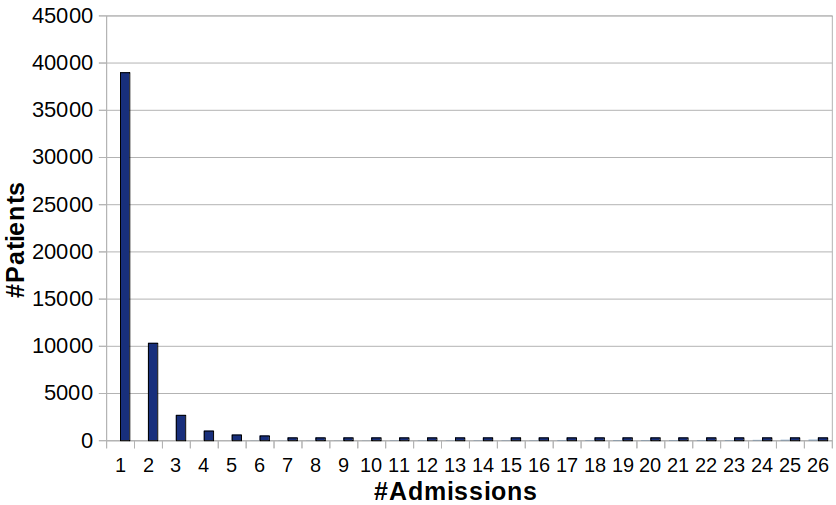}
    \vspace{0.2cm}
    \caption{Distribution of the number of patients with respect to the number of admissions in Mimic-III.}
    \label{fig:admXpat-dist}
\end{figure}

The total number of admissions in the dataset is 58,976. However, the distribution of patients considering the number of admissions is skewed, see Figure~\ref{fig:admXpat-dist} -- 38,983 patients have one single admission. This fact severely reduces the dataset as only data of patients with at least two admissions is useful for trajectory prediction. Besides that, a few admissions do not have any related ICD-9 codes, and others are not meaningful (negative time duration). With these restrictions, the number of admissions falls to 19,911; the number of patients falls from 46,520 to 7,483.

\subsubsection*{ICD-9 encoding}
\begin{figure}[ht!]
\centering
\includegraphics[width=0.5\textwidth]{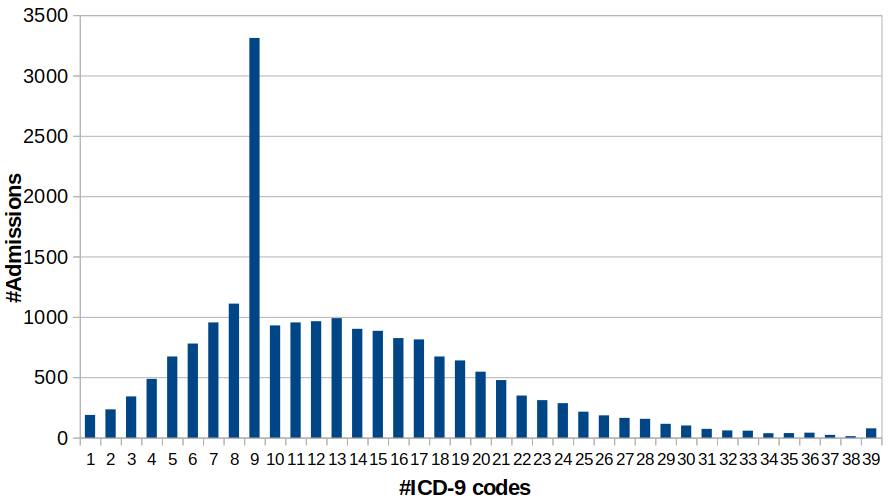}
\vspace{0.2cm}
\caption{Distribution of the number of admissions with respect to the number of diagnosis codes in Mimic-III.}
\label{fig:codesXadms-dist}
\end{figure}

Concerning the encoding of diagnoses, the drawback comes from the high cardinality of the ICD-9 standard, whose number of diagnosis codes sums up to 15,072 (the newer ICD-10 is over 4 times bigger). This cardinality refers to the granularity of details, which describes a disease along with its possible clinical manifestations. In Mimic-III, a total of 6,984 codes appear in the database instance -- Figure \ref{fig:codesXadms-dist} presents the distribution of the number of codes with respect to the number of admissions; although the distribution is not Gaussian due to the outlier of 9 codes per admission, its nearly Gaussian shape allows us to consider the simple average of 13 codes per admission as a reasonable descriptive parameter. As a result, the task of predicting the codes of the next admission lies in the range of $6,984!/6,971!$, or nearly $9.3*10^{49}$ possibilities.

\subsubsection*{Alternative CCS encoding}
The high granularity of the ICD-9 standard is a problem not restricted to this work; rather, it is a recurring problem in several research activities. The Healthcare Cost and Utilization Project (HCUP), a North-American association dedicated to healthcare research has tackled the problem by issuing the Clinical Classifications Software (CCS) encoding~\cite{CCS2015}. Their classification scheme (not a software) defines a specialist-established tabular mapping from ICD-9 to a less granular descriptive standard, the CCS. The goal is to ease statistical analysis and reporting. Table~\ref{tab:ccs_tuberculosis} illustrates the mapping of the disease {\it tuberculosis} from ICD-9 to CSS -- in this example, 426 ICD-9 codes become 1 CCS code. The complete mapping scheme converts 15,072 ICD-9 codes into 285 CCS codes; in the case of Mimic-III (patients with at least two admissions), the mapping corresponds to the use of 271 CCS codes instead of 4,893 ICD-9 codes.

\begin{table*}[ht!]
    \centering
    \begin{tabular}{clcc} \hline
        \textbf{\begin{tabular}[c]{@{}c@{}}ICD-9\\ code\end{tabular}} & \multicolumn{1}{c}{\textbf{\begin{tabular}[c]{@{}c@{}}ICD-9\\ description\end{tabular}}} & \multicolumn{1}{l}{\textbf{\begin{tabular}[c]{@{}l@{}}CCS\\ code\end{tabular}}} & \textbf{\begin{tabular}[c]{@{}c@{}}CCS\\ description\end{tabular}} \\ \hline
        01000 & Prim Tuberculosis Complex-unspec & 1 & Tuberculosis \\ \hline
        01001 & Prim Tuberculosis Complex-no Exam  & 1 & Tuberculosis \\ \hline
        01002 & Prim Tuberculosis Complex-exm Unkn & 1 & Tuberculosis \\ \hline
        \ldots   & \ldots        & 1 & Tuberculosis \\ \hline
        01894 & Miliary Tuberculosis Nos-cult Dx & 1 & Tuberculosis \\ \hline
        01895 & Miliary Tuberculosis Nos-histo Dx  & 1 & Tuberculosis \\ \hline
        01896 & Miliary Tuberculosis Nos-oth Test  & 1 & Tuberculosis \\ \hline
    \end{tabular}
    \vspace{0.2cm}
    \caption{Example of an ICD-9 to CCS mapping; the actual mapping converts 426 possible tuberculosis situations into 1 CCS description.}
    \label{tab:ccs_tuberculosis}
\end{table*}

With the CCS encoding, the problem of predicting the codes of the next admission falls from $9.3*10^{49}$ possibilities to $271!/258!$, or nearly $3.2*10^{31}$ possibilities -- that is, 18 orders of magnitude fewer possibilities. This simplification significantly improves the prediction performance. Of course, the choice for a less granular code has a price; the descriptive results of the predictions are much less detailed; so instead of ``{\it Tuberculosis of ear, tubercle bacilli found (in sputum) by microscopy}'', the diagnosis will state only ``{\it Tuberculosis''}. In the case of Mimic-III, this is a non-avoidable workaround because 19,911 samples are not enough to train an artificial neural network to predict sets of 13 codes, each one pertaining to a 4,893-codes domain.

\subsection{Data and problem modeling}
\label{sec:problem}
\vspace{0.2cm}
The problem treated here is stated as: {\em given a patient's sequence of admissions, possibly stored as an EHR, predict the most probable diagnoses that shall appear in the next admission of this patient at a given time $t+1$.} A patient's admission refers to a pair $a_i=(t_i,\boldsymbol{d_i})$, in which $i$ is the temporal order of the admission, $t_i$ is the timestamp stating when the admission occurred, and $\boldsymbol{d_i}=\{d_{i,0}, d_{i,1}, ..., d_{i,n-1}\}$ is an unordered set of $|n|$ diagnoses codes, so that $d_{i,j} \in D$, in which $D$ is a standard set of codes such as ICD-9 or CCS. Furthermore, each patient's EHR refers to a set of $|m|$ admissions $A=\{a_0, a_1, ..., a_{m-1}\}$. In our problem setting, for any admission $\boldsymbol{a_i}=\boldsymbol{x_{t}}$, we want to predict the codes of admission $\boldsymbol{y_{t+1}}$; the prediction set corresponds to $\boldsymbol{a_{i+1}}=\boldsymbol{\hat{y}_{t+1}}$ -- in the context of artificial neural networks, we want to predict the following probabilities:

\begin{equation}
    \label{eq:probabilities}
    \boldsymbol{\hat{y}_{t+1}}=\boldsymbol{a_{i+1}}=\{P(d_{i+1,j}|\boldsymbol{a}_{0:i})\}
\end{equation}
\noindent{for\ $0 \leq j \leq (|D|-1)$}.
{\ }\\

That is, for each possible code $d_j \in D$, and given admissions $\boldsymbol{a_0}$ through $\boldsymbol{a_i}$, we want to compute its probability of appearing in the next admission $\boldsymbol{a_{i+1}}$. Or, in the conventional notation of the output of an artificial neural network, we seek to compute $\boldsymbol{\hat{y}_{t+1}}$. Notice that, as the outcome is a set of probabilities, it can be interpreted as a set of recommendations.

Considering this problem setting, the input of an admission $\boldsymbol{a_i}$ to an artificial neural network comes in the form of a $|D|$-dimensional multi-hot vector defined, in programming (array-like) notation, as:

\begin{equation}
    \label{eq:hotvector}
    hot\_vector[j]=
    \begin{cases}
      1, & \text{if}\ d_{j} \in \boldsymbol{a_i} \\
      0, & \text{otherwise}
    \end{cases}
\end{equation}
\noindent{for $0 \leq j \leq (|D|-1)$}.
{\ }\\

Since we use batch processing instead of a hot vector per iteration, in practice, Equation \ref{eq:hotvector} expands to a set of $l$ patients $P=\{\boldsymbol{p_0}, \boldsymbol{p_1}, ..., \boldsymbol{p_{l-1}}\}$, each one with a set of $m$ $n$-dimensional admissions $p_h=\{\boldsymbol{a_{h,0}}, \boldsymbol{a_{h,1}}, ..., \boldsymbol{a_{h,m-1}}\}$, which corresponds to the following input tensor:

\begin{equation}
    \label{eq:tensor}
    x[i][h][j]=
    \begin{cases}
      1, & \text{if}\ d_{j} \in a_{h,i} \\
      0, & \text{otherwise}
    \end{cases}
  \end{equation}
\noindent{for $\ 0 \leq i \leq (m-1), 0 \leq h \leq (l-1),\ and\ 0 \leq j \leq (|D|-1)$}.
{\ }\\

In Equation \ref{eq:tensor}, the admissions become the first dimension of the tensor corresponding to slices (orthogonal to its depth axis). This is meant for a simplified flow through the network; it is more convenient to have one patient per line, and one code per column, which makes algebraic operations simpler since the computation is oriented to admissions. Also notice that the patients have different numbers of admissions, and admissions have different numbers of codes; to cope with that, elements with smaller cardinalities are padded with 0's, which demanded the use of masking to prevent residual computations in the padded positions of the tensor.

\subsection{Recurrent neural networks and fine tuning}
\vspace{0.2cm}
We used RNNs, whose principle is to use self-loop connections and a set of information gates whose dynamic produces a memory of past events across time steps -- they contrast with feed-forward-only networks, which do not use self-loops nor memory.
In order to design an architecture with performance superior to existing works, we considered the following types of RNNs: Jordan's network \cite{JORDAN1997471}, Long-Short Term Memory (classic \cite{Hochreiter-1997} and Google's \cite{45379}), Gated Recurrent Units (classic \cite{ChoMBB14} and minimal \cite{Zhou2016}), and DoctorAI (GRU+embedding) \cite{Choi2016-doctorai}; we also considered a feed-forward-only network for comparison and bidirectional recurrent neural networks~\cite{Schuster1997}. We used many auxiliary techniques, including Xavier initialization \cite{pmlr-v9-glorot10a}, dropout, L2 regularization, and addition of Gaussian noise to the input to prevent overfitting; gradient clipping and ADADELTA \cite{abs-1212-5701} for convergence. We considered activation functions Leaky Rectified Linear Unit, sigmoid, hyperbolic tangent, and classical Rectified Linear Unit \cite{Goodfellow-et-al-2016}. Gradient clipping was particularly effective in reducing the loss during each training epoch, although slowing down the convergence.

\section{LIG-Doctor}
\label{sec:LIG-Doctor}
\vspace{0.25cm}
In this section, we introduce our neural network architecture designed to make diagnoses predictions; in the next section, Section \ref{sec:experiments}, we demonstrate that our architecture achieves prediction performance  superior  to  existing  works  in  the  literature.   In  order  to  determine  the  best  neural network architecture, we had to make three decisions: (i) which neural network cell to use – see Section \ref{sec:choosing}; how many layers and neurons to employ – Section \ref{sec:architec}; and (iii) which architecture to use – Section \ref{sec:bidirec}.

\subsection{Choosing an artificial neural network}
\label{sec:choosing}
\vspace{0.25cm}
After extensive testing, the main symptom of our problem setting was its susceptibility to the number of parameters; every time we added a significant number of parameters (layers or neuron nodes), the recall would pointedly fall. We hypothesize that the small number of instances of Mimic-III made it difficult to have the network learn the underlying patterns, therefore reducing the recall for both training and testing. To cope with that, we tested many techniques, as presented in Section \ref{sec:cells}. Among the recurrent networks found in the most-accepted literature, the one with the smallest number of weights is Jordan's network; the one with the biggest number of weights is Google's LSTM. The GRU network demonstrated a higher performance than Jordan's by using more weights, but with less performance than Google's LSTM. The Minimal GRU (MGRU) network uses even fewer weights than the classical GRU and, just as demonstrated by its authors, it did not lose performance despite using fewer gates -- its performance was slightly superior to Google's LSTM and classical GRU, but demanding less processing time. Hence, MGRU was chosen to be the core of our architecture; its equations are:

\begin{subequations}
\label{eq:mgru}
\begin{equation}
\boldsymbol{f_t} = \sigma(\boldsymbol{x_t}W^f + \boldsymbol{h_{t-1}}U^f x_t + \boldsymbol{b^f})
\end{equation}
\begin{equation}
\boldsymbol{\tilde{h_t}} = tanh(\boldsymbol{x_t}W^h + U^h (\boldsymbol{f_t} \odot \boldsymbol{h_{t-1})} + \boldsymbol{b^h})
\end{equation}
\begin{equation}
\boldsymbol{h_t} = (1 - \boldsymbol{f_t}) \odot \boldsymbol{h_{t-1}} + \boldsymbol{f_t} \odot \boldsymbol{\tilde{h_t}}
\end{equation}
\end{subequations}

\noindent{where $\sigma = \frac{e^x}{e^x + 1}$ is the sigmoid activation function; $tanh = \frac{e^x - e^{-x}}{e^x + e^{-x}}$ is the hyperbolic activation function; the $W$'s and $U$'s are the weights to be optimized, together with the biases indicated with $\boldsymbol{b}$'s. We initialize the squared matrices ($U$) using identity, and the other matrices ($W$) using a Gaussian distribution with $zero\ mean$ and $variance = \sqrt{\frac{2}{|input|+|output|}}$. Notice that the order of the dot products depends on the orientation of the input data; we list the equations in the same order as our implementation code to ease understanding -- see Section \ref{sec:problem}.}

\subsection{Determination of the network architecture}
\label{sec:architec}
\vspace{0.25cm}
Here, the first issue was to find an optimal number of neurons to use in the hidden layers. We tested layers ranging from 100 to 3,000 neurons -- to our surprise, we reached a recall plateau at around a number of neurons equal to the number of neurons in the input layer, as demonstrated in Section \ref{sec:cardinal}. Following Equations \ref{eq:hotvector} and \ref{eq:tensor}, this number equals the number of distinct CCS diagnosis codes. For the current Mimic-III dataset instance, this number is 271, as discussed in Section \ref{sec:data}. Similarly, despite the intuition that more layers should lead to better performance, our experiments demonstrated, for every tested technique, that the higher the number of hidden layers the worst the performance. We observed lower performance in both training and testing, so it was not a matter of overfitting, but of capacity to learn the underlying function. This result was intriguing, especially because many authors advocate their technique to be immune to the vanishing/exploding gradient problem or to support deeper networks satisfactorily. We did not further investigate the underlying reasons, but just verified that these claims were not valid for our problem setting -- notwithstanding, this problem has been a topic of active research \cite{PascanuGCB13}. In fact, our results are in sync with a recent work, by Frankle and Carbin \cite{FrankleCargin2019}, who state that neural networks can be as much as 90\% smaller without losing performance.

In our tests, hence, we designed an architecture with one input layer, one MGRU hidden layer, and one standard output layer before the softmax probability distribution. Despite the satisfactory results, we hypothesized that more weights could help because they can detect more about the underlying patterns. However, since stacking more layers did not help, we explored using the principle of bidirectional recurrent neural networks.

\subsection{Adding bi-directional parallelism}
\label{sec:bidirec}
\vspace{0.25cm}
Bidirectional recurrent neural networks connect two processing flows computed in opposite directions in relation to the temporal dimension of the data. As a result, the output layer gets information regarding the past and the future states simultaneously. For our problem setting, this architecture represented significant performance gains, as presented in Section \ref{sec:further}. This design corresponds to having two networks working in parallel instead of only one deeper network; as a result, the design is immune to gradient problems when considering both networks simultaneously -- their parallelism does not produce stacked layers. From an implementation point of view, the backward processing is achieved by simply duplicating the recurrent network and having it fed with data reversed with respect to its first dimension, which is ordered according to the temporal information of the admissions -- see Equation \ref{eq:tensor}. The end of the architecture counts with a feed-forward flow that starts with a joining layer to combine the outcomes of the two networks by means of a weighted sum of the forward and backward computations. Given a forward hidden layer computation $\boldsymbol{h_t^{fwd}}$ and a backward hidden layer computation $\boldsymbol{h_t^{bwd}}$, we obtain their joint weighted sum $\boldsymbol{h_t^{joint}}$ according to:

\begin{figure*}[thb!]
    \centering
    \includegraphics[width=.95\linewidth]{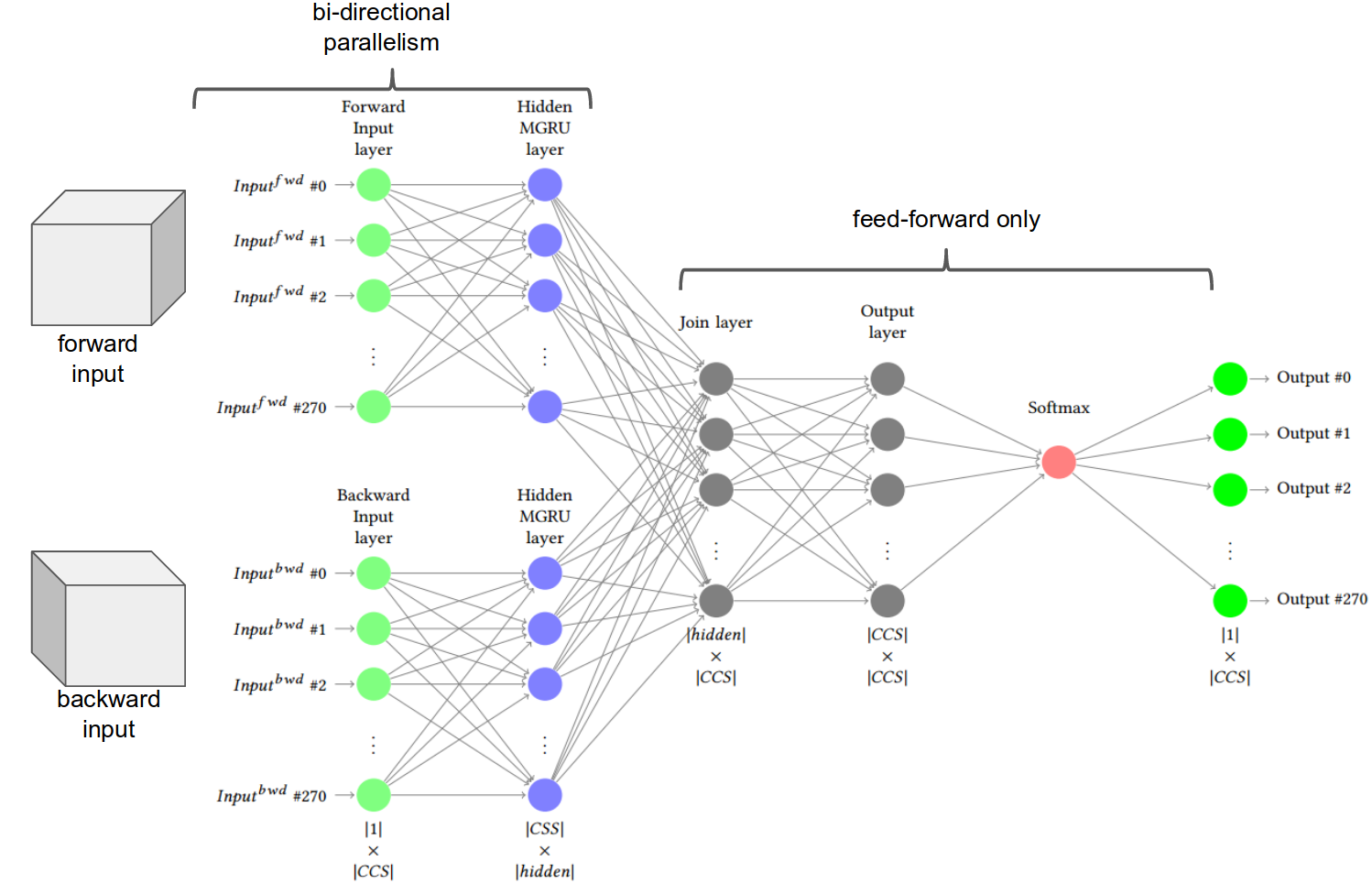}
    \vspace{0.2cm}    
    \caption{Layers diagram of LIG-Doctor -- a bidirectional minimal Gated Recurrent Unit network. We present the cardinalities of the processing matrices below the layers of the network.}
    \label{fig:architecture}
\end{figure*}

\begin{multline}
    \boldsymbol{h_t^{joint}} = LReLU(\\
    \boldsymbol{h_t^{fwd}}V^{fwd} + \boldsymbol{h_t}^{bwd}V^{bwd} + \boldsymbol{b^{joint}},\\
    \alpha^j)
\end{multline}

After that, the final output probabilities come from an output layer that feeds into a softmax operator:

    \begin{multline}
    \boldsymbol{\hat{y}_{t+1}}=softmax(\\
    LReLU(\boldsymbol{h_t^{joint}}W^{out}+\boldsymbol{b^{out}}),\\
    \alpha^o)
    \end{multline}

\noindent{where LReLU corresponds to activation function Leaky Rectified Linear Unit (LReLU) with slopes $\alpha^j$ and $\alpha^o$ as extra optimization parameters introduced in $\boldsymbol{h_t^{joint}}$ and $\boldsymbol{\hat{y}_{t+1}}$, respectively. The use of the parameterized LReLU at the feed-forward stage of the network demonstrated superior results with respect to recall and speed of convergence if compared to functions sigmoid, hyperbolic tangent, and classical Rectified Linear Unit (ReLU).}

Figure \ref{fig:architecture} illustrates the entire architecture, whose goal is to compute the following optimization:

\begin{equation}
    \argmin_{\Theta_{LIGDoc}}(Loss(\boldsymbol{y_{t+1}}, \boldsymbol{\hat{y}_{t+1}}))
\end{equation}

\noindent{where $\Theta_{LIGDoc}=$ $\{W^{ffwd}$, $U^{ffwd}$, $\boldsymbol{b^{ffwd}}$, $W^{hfwd}$, $U^{hfwd}$, $\boldsymbol{b^{hfwd}}$, $W^{ffwd}$, $U^{fbwd}$, $\boldsymbol{b^{fbwd}}$, $W^{hbwd}$, $U^{hbwd}$, $\boldsymbol{b^{hbwd}}$, $V^{fwd}$, $V^{bwd}$, $\boldsymbol{b^{joint}}$, $\alpha^j$, $W^{out}$, $\boldsymbol{b^{out}}$, $\alpha^o$ $\}$ is the set of parameters of the architecture, and the loss function corresponds to the cross entropy function computed over a multi-hot vector of known codes $\boldsymbol{y_{t+1}}$ and a vector of code probabilities $\boldsymbol{\hat{y}_{t+1}}$:}

\begin{equation}
\label{eq:entropy}
\begin{split}
    Loss(\boldsymbol{y_{t+1}}, \boldsymbol{\hat{y}_{t+1}}) = \sum_{j=0}^{|D|-1}(y_{t+1,j}log(\hat{y}_{t+1,j}) +\\ (1-y_{t+1,j})log(1-\hat{y}_{t+1,j}))
\end{split}
\end{equation}

According to the loss function, the closer to $1$ a  probability is, the smaller is the loss.

\section{Experiments}
\label{sec:experiments}
\vspace{0.25cm}
We present evaluation results concerning the multiple techniques found in the literature. The results are meant to justify our design decisions, as well as to guide future researchers in solving similar problems. We also discuss our best results compared to the works presented in Section \ref{sec:relatedworks}.

\subsection{Experimental setup}
\label{sec:setup}
\vspace{0.2cm}
For training and testing, we used 90\% and 10\% of the patients, respectively. The training occurred until the model recorded 10 consecutive epochs without improvement as measured by reductions in the cross-entropy loss -- refer to Equation \ref{eq:entropy}. The code was written over the framework Theano and ran on GPU Nvidia GeForce GTX 1080 Ti; Debian operating system with 256 MB of memory.


\begin{table*}[bht!]
    \centering
        \begin{tabular}{c|c|c|c|c|c|c|c|c|} \cline{2-9}
             & {\cellcolor[rgb]{\tcolor}}Random & \textbf{{\cellcolor[rgb]{\tcolor}}FF-Only} & {\cellcolor[rgb]{\tcolor}}Jordan's & {\cellcolor[rgb]{\tcolor}}DoctorAI & {\cellcolor[rgb]{\tcolor}}LSTM & \textbf{{\cellcolor[rgb]{\tcolor}}LSTM Google} & {\cellcolor[rgb]{\tcolor}}GRU & \textbf{{\cellcolor[rgb]{\tcolor}}Min GRU} \\ \hline
            \multicolumn{1}{|c|}{{\cellcolor[rgb]{\tcolor}}Recall@10} & 0.04 & \textbf{0.46} & 0.43 & 0.44 & 0.46 & \textbf{0.46} & 0.45 & \textbf{0.46} \\ \hline
            \multicolumn{1}{|c|}{{\cellcolor[rgb]{\tcolor}}Recall@20} & 0.07 & \textbf{0.63} & 0.60 & 0.62 & 0.62 & \textbf{0.64} & 0.63 & \textbf{0.64} \\ \hline
            \multicolumn{1}{|c|}{{\cellcolor[rgb]{\tcolor}}Recall@30} & 0.11 & \textbf{0.71} & 0.70 & 0.72 & 0.72 & \textbf{0.73} & 0.73 & \textbf{0.73} \\ \hline
            \multicolumn{1}{|c|}{{\cellcolor[rgb]{\tcolor}}Iterations} & 1 & \textbf{67} & 124 & 47 & 61 & \textbf{38} & 34 & \textbf{38} \\ \hline
            \multicolumn{1}{|c|}{\textbf{{\cellcolor[rgb]{\tcolor}}Time (s)}} & 23.49 & \textbf{53.78} & 86.01 & 52.48 & 103.74 & \textbf{74.37} & 69.25 & \textbf{59.12} \\ \hline
        \end{tabular}
    \vspace{0.2cm}
    \caption{Direct comparison of eight types of artificial neuron cells using metric Recall@k for $k \in \{10, 20, 30\}$. Average values computed after three runs on three different randomized splits of Mimic-III using the settings described in Section \ref{sec:LIG-Doctor}.}
    \label{tab:neuroncells}
\end{table*}

\subsection{Evaluation metrics}
\vspace{0.2cm}
Due to the characteristics of the problem, we employ a metric commonly used for recommendation systems: recall at top-k recommendations. In our case, the top $k$ recommendations refer to the $k$ diagnosis codes in $\boldsymbol{\hat{y}_{t+1}}$ that have the highest probabilities -- see Equation \ref{eq:probabilities}. Considering the top-k recommendations, Recall@k refers to the percentage (ratio) of recommended codes that are correct (actually relevant), expressed by $recall@k=\frac{\#correctly\ recommended\ codes}{|\boldsymbol{y_{t+1}}|}$ -- a code is correct if it pertains to the answer set $\boldsymbol{y_{t+1}}$.

\subsection{Direct comparison of neuron cells}
\label{sec:cells}
\vspace{0.2cm}
This first round of experiments refers to Section \ref{sec:choosing}; here we compared eight techniques with respect to their Recall@k: random-initialization-only without training, feed-forward-only without recurrent cells, Jordan's network, DoctorAI, LSTM, Google LSTM, GRU, and Minimal GRU. We executed each technique with exactly the same hyperparameters and fine-tunings used in our methodology (see Sections \ref{sec:LIG-Doctor} and \ref{sec:setup}) -- actually, for each experiment, we simply changed the cell type, keeping everything else the same. Each technique ran over three randomized versions of Mimic-III split in 90\% for training and 10\% reserved for testing. The average results, presented in Table \ref{tab:neuroncells}, demonstrate superior performance, with up to 73\% accuracy for Recall@30, for techniques Google LSTM and Minimal GRU, which also had the smallest number of iterations before convergence. Surprisingly, the feed-forward-only network had a performance comparable to classic LSTM and GRU, with up to 72\% accuracy for Recall@30. The probable reason is that a great portion of the patients have only two admissions, case when the time-awareness or recurrent networks is not necessary. After these results, we chose techniques feed-forward-only, Google LSTM and Minimal GRU for further investigation.

\subsection{Experimenting with cardinalities}
\label{sec:cardinal}
\vspace{0.2cm}
After choosing the most adequate neuron cells, we proceeded with empirical tests regarding the number of layers and number of neurons, as explained in Section \ref{sec:architec}. For each kind of cell, we experimented with up to 3 layers and with 271, 542, and 1,084 cells, in a total of 9 different settings for each cell type -- the number of cells is a multiple of the size of the input layer, as explained in Section \ref{sec:LIG-Doctor}. In Table \ref{tab:cardinalities}, we see that more layers caused the system to lose performance - meanwhile the number of neurons did not affect the results so much, but, of course, it demanded more processing time. The feed-forward-only network was the least-resilient setting; its performance decreased from 71\% at one 271-nodes layer to 55\% for three 1,084-nodes layer. Google LSTM and Minimal GRU, again, had similar performances -- their Recall@30 ranged from 73\% to nearly 69\%, with Minimal GRU presenting slightly better results. Concerning the processing time, since Minimal GRU has fewer gates, it computes faster than LSTM, even when it runs for a few more iterations. Considering all these aspects, we decided for Minimal GRU as the neuron cell of our architecture, and, also, for one single 271-nodes hidden layer.


\begin{table}[hbt!]
    \centering
    \resizebox{.95\columnwidth}{!}{%
        \begin{tabular}{|c|c|c|c|} 
            \cline{2-4}
            \multicolumn{1}{c|}{}                   & \multicolumn{3}{c|}{{\cellcolor[rgb]{\tcolor}}1 Layer}   \\ \hline
            \rowcolor[rgb]{\tcolor} Nodes & Minimal GRU & Google's LSTM & Feed-forward-only \\ \hline
            {\cellcolor[rgb]{\tcolor}271} & 0.73        & 0.73          & 0.71 \\ \hline
            {\cellcolor[rgb]{\tcolor}542} & 0.74        & 0.73          & 0.72 \\ \hline
            {\cellcolor[rgb]{\tcolor}1,084} & 0.73        & 0.72          & 0.72 \\ \hline
            \multicolumn{1}{c|}{}                   & \multicolumn{3}{c|}{{\cellcolor[rgb]{\tcolor}}2 Layers}  \\ \hline
            \rowcolor[rgb]{\tcolor} Nodes & Minimal GRU & Google's LSTM & Feed-forward-only \\ \hline
            {\cellcolor[rgb]{\tcolor}271} & 0.69        & 0.69          & 0.65 \\ \hline
            {\cellcolor[rgb]{\tcolor}542} & 0.71        & 0.71          & 0.65 \\ \hline
            {\cellcolor[rgb]{\tcolor}1,084} & 0.71        & 0.72          & 0.64 \\ \hline
            \multicolumn{1}{c|}{}                   & \multicolumn{3}{c|}{{\cellcolor[rgb]{\tcolor}}3 Layers}  \\ \hline
            \rowcolor[rgb]{\tcolor} Nodes & Minimal GRU & Google's LSTM & Feed-forward-only \\ \hline
            {\cellcolor[rgb]{\tcolor}271} & 0.67        & 0.65          & 0.60 \\ \hline
            {\cellcolor[rgb]{\tcolor}542} & 0.69        & 0.67          & 0.57 \\ \hline
            {\cellcolor[rgb]{\tcolor}1,084} & 0.69        & 0.68          & 0.55 \\ \hline
        \end{tabular}
    }
    \vspace{0.3cm}
    \caption{Performance comparison of neuron networks Feed-forward-only, Google's LSTM, and Minimal GRU with the number of layers ranging from 1 to 3, each with 271, 542, and 1,084 nodes.}
    \label{tab:cardinalities}
\end{table}

\begin{table*}[bht!]
    \centering
        \begin{tabular}{c|c|c|c|c|c|c|c|} \cline{2-8}
            & {\cellcolor[rgb]{\tcolor}}Bi-directional & {\cellcolor[rgb]{\tcolor}}+embedding & {\cellcolor[rgb]{\tcolor}}+interval & {\cellcolor[rgb]{\tcolor}}+duration & {\cellcolor[rgb]{\tcolor}}+type & {\cellcolor[rgb]{\tcolor}}+interval+duration & {\cellcolor[rgb]{\tcolor}}+interval+duration+type \\ \hline
            \multicolumn{1}{|c|}{{\cellcolor[rgb]{\tcolor}}Recall@10} & 0.51 & 0.50 & 0.51 & 0.52 & 0.51 & 0.52 & 0.51 \\ \hline
            \multicolumn{1}{|c|}{{\cellcolor[rgb]{\tcolor}}Recall@20} & 0.68 & 0.66 & 0.68 & 0.68 & 0.68 & 0.68 & 0.68 \\ \hline
            \multicolumn{1}{|c|}{{\cellcolor[rgb]{\tcolor}}Recall@30} & 0.76 & 0.75 & 0.76 & 0.76 & 0.76 & 0.76 & 0.76 \\ \hline
            \multicolumn{1}{|c|}{{\cellcolor[rgb]{\tcolor}}Iterations} & 41 & 32 & 50 & 42 & 41 & 47 & 42 \\ \hline
        \end{tabular}
    \vspace{0.25cm}
    \caption{Performance of our methodology when combined with further techniques and data features. The best results came with the use of the duration of the hospital admission concatenated to the input data (no embedding layer).}
    \label{tab:advanced}
\end{table*}

\subsection{Further design improvements}
\label{sec:further}
\vspace{0.3cm}
After deciding for the Minimal GRU cell, and for the cardinality of neurons and layers. the next step was to use more elaborate techniques to further improve the performance. We experimented with the principle of bidirectional recurrent neural networks, which led us to the parallel architecture discussed in Section \ref{sec:bidirec} and to a higher performance improvement -- refer to the first column of Table \ref{tab:advanced}. For the bi-directional Minimal GRU network, the Recall ranged from 51\% at $k=10$ to 76\% at $k=30$, more than 10\% better than any other setting. Over this architecture, we experimented other techniques, including the use of an embedding layer before the hidden layers; and the use of extra features (duration of admission, interval between admissions, and type of the admission).

As presented in Table \ref{tab:advanced}, the embedding layer just reduced the performance; a side effect that we verified for all the settings previously reported -- probably, the smaller cardinality of the CCS encoding does not sustain the use of embedding. The use of unsupervised pre-training, which answers for a more adequate initialization of the weights based on auto-encoding-like preprocessing, was capable of reducing the time of convergence; however, we verified no significant performance improvements. Finally, the use of extra features found in the database, namely the type of the admission (newborn, elective, emergency, or urgent), the interval between admissions, and the duration of the admissions, provided slight improvements -- refer to columns 3 to 7 in Table \ref{tab:advanced}. We used these extra features via concatenation to the input tensor, without an embedding layer -- the type in the form of a 4-codes hot-vector; the time in the form of a single normalized extra slice. The first of these features, type, is very particular to the Mimic-III; the duration, however, applies to any other medical dataset -- provided that anonymization did not corrupt the timings. The final highest Recall, achieved with duration, ranged from 51\% at $k=10$ to 76\% at $k=30$.

\subsection{Comparison to related works}
\vspace{0.3cm}
The work of Pham et al. \cite{PHAM2017218} recommended both the use of an embedding layer and LSTM cell; our results, though, demonstrated that this is not the case for all settings -- embedding, in particular, was a very bad design choice. We directly compared to the methodology of Choi et al. \cite{Choi2016-doctorai}, with exactly the same settings and using the code provided by the authors -- we report better results, as presented in Table \ref{tab:neuroncells}. In a broad sense, these two former works are narrow with respect to their domain, dealing with very specific diseases. With respect to previous works that do not rely on artificial neural networks -- mentioned in Section \ref{sec:relatedworks}, it is possible to affirm that they do not straightly adapt to different settings, requiring very specialized data and problem modeling; or, they demonstrate performances that demand million-scale volumes of records. Meanwhile, Deep Learning has become one of the most active areas of research; improvements appear every day, improving existing architectures or introducing hyper-parameters that render better performance. Our methodology relates to all the aforementioned issues.

Furthermore, it is worth to mention that a strict consolidated benchmark for patient trajectory prediction is not yet of broad use. The field still has much to evolve in order to ultimately evaluate the performance of one given predictor. The Mimic-III dataset is an open initiative to fill this benchmark gap; the research community shall benefit from future works that experiment on Mimic-III beyond their private datasets.

\section{Conclusions}
\vspace{0.3cm}
We conducted broad experimentation over the Mimic-III dataset, provided by MIT. The experimentation considered a vast set of techniques concerning sequence to sequence prediction, dealing with recurrent neural networks, and other related methods. While searching for the best design, we had interesting insights that might inspire further work and/or guide design processes of similar problems. Our first finding was that recurrent network techniques do not accomplish many of the claims that abound in the respective literature. Specifically, we verified that they do not support the stacking of layers -- the more layers, the worse the performance; effectively, our design ended up with one single layer. We also noticed that the networks whose cells used more gates, like LSTM and GRU, had the same performance as of the network based on the much simpler Minimal GRU cells, which we ended up choosing for our design. When one considers the theoretical claims on why each gate is part of a given network, the facts do not support the theoretical premises -- for our specific settings, the fewer gates, the better. Complexity seemed no to be the path to follow. These two findings were not exhaustively investigated, notwithstanding, our results alert that some assumptions taken for granted must be revisited.

Concerning the Mimic-III dataset, we found that, although it can support the training of a neural network aiming at medical prognosis, this is feasible only by using a less granular coding for diseases, as the HCUP-CCS encoding that we used instead. In fact, we verified that the ICD-9 is way too vast for a dataset with the size of Mimic-III; moreover, if we consider that the newer ICD-10 encoding is over 4 times more granular, the research community shall consider that, although it is adequate for precise medical description, it might not be suitable for effective statistical and analytical tasks, raising a demand for alternative database projects.

Finally, research lines can be further investigated as a continuation of this work. The success in using a parallel architecture demonstrated to be a promising design decision; even further, this design can be extrapolated to more than two parallel networks, each one benefiting from different characteristics of the data -- actually, Mimic-III has many more semantic features that shall support further investigation. We also suggest that the whole methodology be experimented over more specific datasets, as for predicting finer onsets, like heart failure, or strokes; and also, for predicting when the next onset might take place, as the data is rich with respect to temporal information. Lastly, the research on adversarial networks for data augmentation appeared shortly after we started working with longitudinal medical data -- this tends to be a topic of active research in the next years.

\section*{Acknowledgments}
\vspace{0.3cm}
This research was financed by Brazilian agencies Coordenacao de Aperfeicoamento de Pessoal de Nivel Superior (CAPES, Finance Code 001); Fundacao de Amparo a Pesquisa do Estado de Sao Paulo (Fapesp, grants 2019/04461-9, 2018/17620-5, 2017/08376-0, 2016/17078-0, and 2019/04461-9); and Conselho Nacional de Desenvolvimento Cientifico e Tecnologico (CNPq, grants 167967/2017-7, and 305580/2017-5). We also thank Nvidia Corporation for donating the GPUs that supported this work.


\input{main.bbl}

\end{document}